# 5

# I'm Sorry to Say, But Your Understanding of Image Processing Fundamentals Is Absolutely Wrong

Emanuel Diamant
*VIDIA-mant, Kiriat Ono*
*Israel*

## 1. Introduction

Among the five human senses through which we explore our surrounding, vision takes a unique and a remarkable place. The lion part of information about our near, medium, and distant environment comes to us via the vision channel. It is, therefore, not surprising that almost a half of our cortex is devoted to visual information processing (Milner & Goodale, 1998). In the course of millions of years of evolution, we have even developed a very special attitude to it – we feel an everlasting "hunger" for new visual information. We are "Infovores", as Irving Biederman (Biederman & Vessel, 2006), one of the founders of the contemporary vision theory, wittily defined.

Maybe, this perpetual yearning is the incentive that made us so inclined to various forms of visual information gathering and accumulation. The story about explosive expansion of camera phones may be a good example here: At the end of the year 2007, Nokia manage to sell almost 440 million mobile phones (obviously, each one equipped with a tiny video camera) which accounted for 40% of all global mobile phone sales (Nokia, 2008). That means, more than a billion mobile phones have been soled worldwide only in one last year!

By the late 2009, the total number of camera phones will exceed that of both conventional and digital cameras shipped since the invention of photography (Thevenin et al., 2008). The result is – an unprecedented and previously unknown flood of visual information in our environment. According to a leading market research firm, Internet video consumption has increased by nearly 100% over the past year: from an average of 700 terabytes/day in 2006, to 1200 terabytes/day in 2007. Internet video uploads have reached 500K uploads/day in 2007 and will grow to 4800K in 2011 (Mobile video, 2008).

That places an urgent demand for a new and previously unknown way of visual information flow handling and management. Certainly, it must be human-like and human-compatible, because Human Visual System (HVS) is the sole information processing system we know that is capable to cope with such problems.

However, by saying that we immediately fall into a trap – we don't know how HVS so perfectly performs its duties. What we do know is that video data sampled by 126 millions of photoreceptors at the eye's retina is immediately converted (as long as the visual input propagates from the eyes to the higher brain processing levels) into meaningful disjointed visual objects, of various complexities. It must be stressed again and again – we do not know how this semantic segmentation is accomplished. But we certainly know that the bulk of visual processing accomplished in the human's brain is performed at the semantic information processing level. Artificial visual systems that we have tirelessly attempted to construct over the last half of a century have always lacked such an ability. The bulk of visual processing carried out in artificial visual systems is constrained to visual data processing only: Pure, exhaustive data processing and nothing more than that.

The apparent difference and incompatibility between these two image processing modalities – pure low-level data processing in human-made visual systems and enigmatic high-level semantic information processing in natural human visual systems – is often overlooked and commonly misinterpreted in the computer vision community. This leads to many funny things that are ubiquitous in computer vision design practice, but they seem far less funny when the production scale of such lapses is regarded. Here are some examples:

The perceptual quality of an image is usually strictly tied with image primary resolution. More pixels in a frame – more valued is the image. Undeniably, this philosophy is the driving force behind the race for megapixel-large image sensors for portable phone cameras, or the High Definition Television Standard for stationary devices. In each case, image high resolution is directly associated with an extremely high volume of raw image data.

Communication bandwidth constraints, power-on-hand limits and other design restrictions request effective signal compression techniques to be used in such data abundant cases. Indeed, carefully designed and skillfully adjusted compression/decompression (encoding/decoding) techniques are generally implemented. Their prime and single purpose: to reduce the data-handling burden. But in the end, the compressed/ decompressed image data would be always again presented to a human observer for final treatment and semantic information processing.

A smart design approach would attempt from the very beginning to encode the semantic objects buried in the image data and to deliver only them to the human disposal. That is exactly what the MPEG-4 Standard designers have in mind when they have introduced the standard's innovative features: VO (Visual Object), VOP (Visual Object Plane), VOL (Visual Object Level). That happened in the year 1994 (Puri & Eleftheriadis, 1998), and expectations for the new video code were very high.

However, as the time passed, nothing has come about in the field. And for a very simple and sad reason: visual object is a semantic entity, which cannot be attained by data manipulations. Standard designers were aware of this problem, and for this reason nothing was said about the way the visual objects have to be discovered and delineated. Hence, all further improvements and modifications the standard went through (and there were a lot of them, the last version of the standard is even named differently – H.264 or MPEG-4 Advanced Video Coding (H.264/AVC)) are concerned only with data coding improvements (Sullivan & Wiegand, 2005).

The consequences of this are easily imaginable: for stationary environments where power dissipation, processing speed limitations and cost restrictions are not a concern, extremely powerful DSPs (Digital Signal Processors) like Analog Devices TigerSHARC ADSP-TS201S with 3.6 GFLOPs processing power are put into work. For those who are not satisfied with such a might – BittWare offers a PCI Mezzanine Card featuring four TigerSHARCs on a single board with a general processing power of 57 GFLOPs (Bittware, 2007).

For the mobile applications, where the restrictions are stern and fixed, the only possible solution is to compromise on image resolution (size). While the sensor resolution has

steadily grown from 1.5 Megapixels (1280x1024) to 5 (2580x1930), 8 (3264x2444), 12 (4220x2820), and at last 14 Megapixels (4570x3050), the actually operated camera-phone images were of the size 80x60 pixels, or 160x120, or finally 352x288 pixels, which is the CIF (Common Intermediate Format) Standard. That is all what the infovore people can get in the real life circumstances.

Another field where vision technology is extensively used is video surveillance. Spurred by increasing public and private security concerns (especially after 9/11), video surveillance systems market is observing an unprecedented growth and expansion. Global video surveillance camera revenue is forecast to grow from $4.9 billion in 2006 to more than $9 billion by 2011 (Video surveillance, 2007).

The general stance is that the driving force behind this expansion is the networked Internet Protocol (IP) video surveillance cameras and IP video servers. Indeed, the IP technology provides the basis for a great leap in video surveillance systems design. However, it has a serious drawback: in terms of useful image resolution the 352x288 CIF standard is the predominating one. From the standpoint of a surveillance system user, the quality of an image in such a system is very dubious. But not this peculiar feature is now in the focus of our concern – visual surveillance implies that the delivered picture is examined and analyzed for scene changes and suspicious event developments (to be detected) and appropriated countermeasures triggered in response. As it was already explained above, this is a sheer semantic information-processing task that only a human being can perform, and none of the existing video surveillance systems can cope with such a task autonomously. What fallows from this, is that a human observer must be attached to the system's display forever: 24 hours a day/7 days a week/52 weeks a year. Otherwise the system is ineffective and useless. However, for such monotonous and boring work humans are the worst candidates. But who cares? To save on expenses, the observer's display usually contains not a single camera output, but is shared between 4, 8, and even 16 camera outputs. The effectiveness of such surveillance systems is less than illusive. The arrogant indifference to human/machine disparities in visual stuff handling is celebrating again. And the market - keeps on growing continuously, every time more and more.

## 2. Tears do not solve problems

The urgent need for machine-based visual systems, which are capable of processing visual information in a human-like intelligent manner, is well understood and widely acknowledged today. Impressive research programs that European Commission runs under its auspice are a good example for this understanding. But the scale of the efforts and billions of Euro put into the enterprise (European IST Research, 2006), cannot explain the lack of the progress we witness during all phases of the projects development. We are now in the 7th Framework Programme (FP7), but nothing serious has happen, and the things seemed to be stalled in a dead-end alley.

That is a proper moment to check again the basic principles we adhere to when we are pursuing our routine research goals. Since we are aimed on human-like visual information processing, we first have to scrutinize the available knowledge about the HVS performance and then to analyse how this knowledge is used in modelling various human-like image-processing tasks.

The classical paradigm of human visual information processing has been established few decades ago by the seminal works of David Marr (Marr, 1978; Marr, 1982), Anne Treisman (Treisman & Gelade, 1980), Irving Biederman (Biederman, 1987), and a large group of their associates and followers. Treisman's "Feature-integration theory" (Treisman & Gelade, 1980) is considered as the most fitting incarnation of the idea. It regards human visual information processing as an interplay of two inversely directed processing streams. One is an unsupervised, bottom-up directed process of initial image information pieces discovery and localization. The other is a supervised, top-down directed process, which conveys the rules and the knowledge that guide the linking and binding of these disjoint information pieces into perceptually meaningful image objects.

Essentially, as an idea, this conception was not entirely new. About two hundred years ago, Kant had depicted the "faculty of (visual) apperception" as a "synthesis" of two constituents: the raw sensory data and the cognitive "faculty of reason" (Hanna, 2004). A century later, Herman Ludwig Ferdinand von Helmholtz (the first who scientifically investigated our senses) had reinforced this view, positing that sensory input and perceptual inferences are different, yet inseparable, faculties of human vision (Gregory, 1979). The novelty of the modern approach was in an introduction of a new concept used for the idea clarification - "visual information" (Marr, 1978). However, a suitable definition of the term was not provided, and the mainstream of relevant biological research has continued (and continues today) to investigate the puzzling duality of the phenomenon by capitalizing on traditional vague definitions of the matters: local and global image content, perceptual and cognitive image processing, low-level computer-derived image features versus high-level human-derived image semantics (Barsalou, 1999; Palmeri & Gauthier, 2004). Putting aside the terminology, the main problem of human visual information processing remains the same: in order to fulfill the intuitively effortless low-level information pieces agglomeration into meaningful semantic objects, the system has to be provided with some high-level knowledge about the rules of this agglomeration. Needless to say, such rules are usually not available. In biological vision research, this dilemma is known as the "binding problem". Its importance was recognized at very early stages of vision research, and massive efforts have been directed into it in order to reach a suitable and an acceptable solution. Despite the continuous efforts, any discernable success has not been achieved yet. (For more details, see Treisman (1996) and the special issue of Neuron (vol. 24, 1999), entirely devoted to this problem).

Unable to reach the required high-level processing (binding) rules, vision research took steps in a forbidden, but possibly an appealing and an enticing direction – to try to derive the needed high-level knowledge from the available low-level information pieces. A rank of theoretical and experimental work has been done in order to support and justify this just-mentioned shift in research aspirations. Two approaches could be distinguished in this regard: chaotic attractor modeling approach (McRae, 2004; Johanson & Lansner, 2006), and saliency attention map modeling approach (Treue, 2003; Itti, 2005). There is no need to review the details of these approaches here. I will only make a note that both of them presume low-level bottom-up processing as the most proper way for high-level information recovery. Both are computationally expensive. Both definitely violate the basic assumption about the leading role of high-level knowledge in the low-level information processing.

In computer vision, the situation is even more bizarre. In fact, computer vision community is so busy with its everyday problems that there is no time to raise basic research ventures. Principal ideas (and their possible solutions) are usually borrowed from biological vision research. Therefore, following the trends in biological vision, the computer vision R&D for decades has been deeply involved in bottom-up pixel-oriented image processing. Low-level image computations have become its prime and persistent goal, while the complicated issues of high-level processing were just neglected and disregarded.

However, it is impossible to ignore them completely. It is generally acknowledged that any kind of image processing is unfeasible without incorporation into it the high-level knowledge ingredients. For this reason, the whole history of computer-based image processing is an endless saga on attempts to seize the needed knowledge in any possible way. The oldest and the most common ploy is to capitalize on the expert domain knowledge and adapt it to each and every application case. It is not surprising, therefore, that the whole realm of image processing has been (and continues to be) fragmented (segmented) according to high-level knowledge competence of the domain experts. That is why we have today: medical imaging, aerospace imaging, infrared, biologic, underwater, geophysics, remote sensing, microscopy, radar, biomedical, X-ray, and so on "imagings".

The advent of the Internet, with huge volumes of visual information scattered over the web, has demolished the long-lasting custom of capitalizing on the expert knowledge. Image information content on the Web is unpredictable and diversified. It is useless to apply specific expert knowledge to a random set of distant images. To meet the challenge, the computer vision community has undertaken an enterprise to develop appropriate (so-called) Content-Based Image Retrieval (CBIR) technologies (Lew et al. 2006). However, deprived of any reasonable sources of the desired high-level information, computer vision designers were forced to proceed in the only one possible direction – trying to derive the high-level knowledge from the available low-level information pieces (Mojsilovic & Rogowitz, 2001; Zhang & Chen, 2003).

It will be a mistake to say that computer vision people are not aware of these discrepancies. On the contrary, they are well informed about what is going on in the field. However, they are trying to justify their attempts by promoting a concept of a "semantic gap", an imaginary gap between low- and high-level image features. They sincerely believe that some day they would be able to bridge over it (Hare et al., 2006).

It is worth to mention that all these developments (feature binding in biological vision and semantic gap bridging in computer vision) are evolving in an atmosphere of total indifference towards preceding claims about high-level information superiority in the general course of visual information processing. Such indifference seems to stem from a very loose understanding about what is the concept of "information", what is the right way to use it properly, and what information treatment options could arise from this understanding.

## 3. Trying to define "What is information?"

I was very proud of myself when it has become clear to me that the problem image processing is subjected to stems from misunderstanding and confusing the duties that machine vision and human vision systems are destined to perform: machine vision systems are for data processing, human vision systems – for information processing. It was clear to me that data and information are different things, and therefore a careless blending of them is harmful and counterproductive (as it follows from the examples provided above). However, my conjectures have not been readily welcomed. My paper submitted to BMCV 2002 Conference was rejected, and the reviewer was very strict in his comments: "The distinction between information and data processing is superficial – you have to be more specific (after all, data is information, isn't it?)".

I was hurt by what has seemed to me as reviewer's ignorance. But later I was forced to learn that that is a well-established, widespread and quite common view on the matters. Luciano Floridi's papers (Floridi, 2003; Floridi 2005; Floridi 2007) are busy with refining "the Standard Definition of semantic information as meaningful data" (!!!). Alas, you cannot quarrel with Floridi. Especially, as your own definition is so vague and muddle-headed that it is better for you to take a stance that "information" is an indefinable entity, like "time" or "space" in classical physics. (Later I have found out that a similar stance is taken by Aaron Sloman (Sloman, 2006) when he compares the indefinable notion of "information" with the indefinable notion of "energy").

Following my own intuition, I have finally hit on something I was so desperately looking for – an information definition fitting my image processing requirements. It turns out that this definition can be derived from Solomonoff's theory of Inference (Solomonoff, 1997), Chaitin's Algorithmic Information theory (Chaitin, 1977), and Kolmogorov's Complexity theory (Kolmogorov, 1965). The results of my investigation have been already published on several occasions, (Diamant, 2003; Diamant, 2004; Diamant, 2005; Diamant, 2007), and interested readers can easily get them from a number of freely accessible repositories (e.g., arXiv, CiteSeer (the former Research Index), Eprintweb, etc.). Therefore, I will only repeat here some important points of these early publications, which properly reflect my current understanding of the matters.

The main point is that **information is a description**, a certain alphabet-based or language-based description, which Kolmogorov's theory regards as a program that, being executed, trustworthy reproduces the original object (Vitany, 2006). In an image, such objects are visible data structures from which an image consists of. So, a set of reproducible descriptions of image data structures is the information contained in an image.

The Kolmogorov's theory prescribes the way in which such descriptions must be created: at first, the most simplified and generalized structure must be described. (Recall the Occam's Razor principle). Then, as the level of generalization is gradually decreased, more and more fine-grained image details (structures) become revealed and depicted. This is the second important point, which follows from the theory's pure mathematical considerations: image **information is a hierarchy of recursive decreasing level descriptions** of information details, which unfolds in a coarse-to-fine top-down manner. (Attention, please: any bottom-up processing is not mentioned here. There is no low-level feature gathering and no feature binding!!! The only proper way for image information elicitation is a top-down coarse-to-fine way of image processing.)

The third prominent point, which immediately pops-up from the two just mentioned above, is that the top-down manner of image **information elicitation does not require incorporation of any high-level knowledge** for its successful accomplishment. It is totally free from any high-level guiding rules and inspirations. That is why I call it **Physical Information** – information that is totally independent of any high level interpretation of it.

What immediately follows from this is that high-level image semantics is not an integrated part of image information content (as it is traditionally assumed). It cannot be seen more as a natural property of an image. Image semantics, therefore, must be seen as a property of a human observer that watches and scrutinizes an image. That is why we can definitely say:

**semantics is assigned to an image by a human observer**. That is strongly at variance with the contemporary views on the concept of semantic information.

Following the new information elicitation rules, it is impossible to continue to pretend that semantics can be **extracted from an image**, (as for example in (Naphade & Huang, 2002)), or should be **derived from low-level information features** (as in (Zhang & Chen, 2003; Mojsilovic & Rogowitz, 2001), and many other analogous publications). That simply does not hold any more.

## 4. Reification of the proposed idea

The new definition of information has forced us to reconsider the traditional way of doing things in image processing. The inevitable change in design philosophy, the validity of new assumptions, the consequences that acceptance of new assumptions imply, all this has motivated us to test the proposed novelties in a framework of visual robot design enterprise – an enterprise, which is aimed on creating an artificial vision system with some human-like cognitive capabilities.

As follows from the preceding discussion, the proposed arrangement must be comprised of two separate loosely coupled parts: Physical Information processing part and Semantic Information processing part. The proposed block-scheme of this arrangement is depicted in Fig. 1.

### 4.1 Physical information processing

The purpose of the Physical Information processing part is to extract the physical information buried in the image data. That is, to provide a description of discernable image data structures present in a given image. In simple words, to provide an initial segmentation of the input image. Afterwards the segmented pieces would be submitted to a process of image analysis and interpretation (in terms of our approach – Semantic Information would be assigned to the input image).

As one can see, the proposed Physical Information processing part is comprised of three sub-units: the bottom-up processing path, the top-down processing path and a stack where the discovered information content (the generated descriptions of it) are actually accumulated. (More details about Physical Information processing can be found in (Diamant, 2004; Diamant, 2005; Diamant, 2005a).

As follows from the early-defined information processing principles (which prescribe that the most general and simplified descriptions have to be derived first), the purpose of the bottom-up processing path is to provide a simplified (compressed, squeezed) copy of an input image. The original image is squeezed along this path to a small size of approximately 100 pixels. The rules of this shrinking operation are very simple and fast: four non-overlapping neighbor pixels in an image at level $L$ are averaged and the result is assigned to a pixel in a higher ($L$+1)-level image, (a so-called 4 to 1 image compression). At the top of the shrinking pyramid, the image is segmented, and each segmented region is labeled. Since the image size at the top is significantly reduced and since in course of the bottom-up image squeezing a severe data averaging is attained, the image segmentation/classification procedure does not demand special computational efforts.

From this point on, the top-down processing path is commenced. At each level, the segmentation maps (intensity and region labels) are expanded to the size of an image at the nearest lower level, (a 1 to 4 expansion). Since the regions at different hierarchical levels do not exhibit significant changes in their characteristic intensity, the majority of newly assigned pixels are determined in a sufficiently correct manner. Only pixels at region borders and seeds of newly emerging regions may significantly deviate from the assigned values. Taking the corresponding current-level image as a reference (the left-side unsegmented image), these pixels can be easily detected and subjected to a refinement cycle. The region labels map is corrected accordingly. In such a manner, the process is subsequently repeated at all descending levels until the segmentation of the original input image is successfully accomplished.

At each processing level, every segmented image object-region (whether just recovered or an inherited one) is registered in the objects' appearance list (the Stocked Level Descriptions rectangle in Fig. 1), which is the third constituting part of the proposed scheme.

The registered object parameters are the available simplified object's attributes, such as size, center-of-mass position, average object intensity and hierarchical and topological relationship within and between the objects ("sub-part of…", "at the left of…", etc.). They are sparse, general, and yet specific enough to capture the object's characteristic features in a variety of descriptive forms.

This way, a practical algorithm based on the announced above principles has been developed and subjected to some systematic evaluations. The results were published, and can be found in (Diamant, 2004; Diamant, 2005; Diamant, 2005a). There is no need to repeat again and again that excellent, previously unattainable segmentation results have been attained in these tests, undoubtedly corroborating the new information processing principles. Not only an unsupervised segmentation of image content has been achieved, (in a top-down coarse-to-fine processing manner, without any involvement of high-level knowledge), a hierarchy of descriptions for each and every segmented lot (segmented sub-object) has been achieved as well. It contains a set object related parameters, which enable subsequent object reconstruction. That is exactly what we have previously defined as **information**. That is the reason why we specify this information as "physical information", because that is the only information present in an image, and therefore **the only information that can be extracted from an image**.

### 4.2 Semantic information processing

Semantic information, which (as we understand now) conveys the property of an external observer, is completely dissociated from the physical information contained in an image. Therefore it must be treated (or modeled) in accordance with observer-specific (his/her) cognitive information processing rules.

What are these rules? A consensus view on this topic does not exist as yet in the biological vision theories as well as in the computer vision practice. So, we have to blaze our own trails. We decided, thus, to meet this challenge by suggesting a new approach based on our previously declared information elicitation principles. The preliminary results of our first attempt have been published elsewhere (Diamant, 2006). As in the case of physical information, we will not repeat here all the details of this publication. Possible implementation details of the Semantic Information processing part (solution) are depicted in Fig. 1. Here we will proceed only with a brief explanation of some of them.

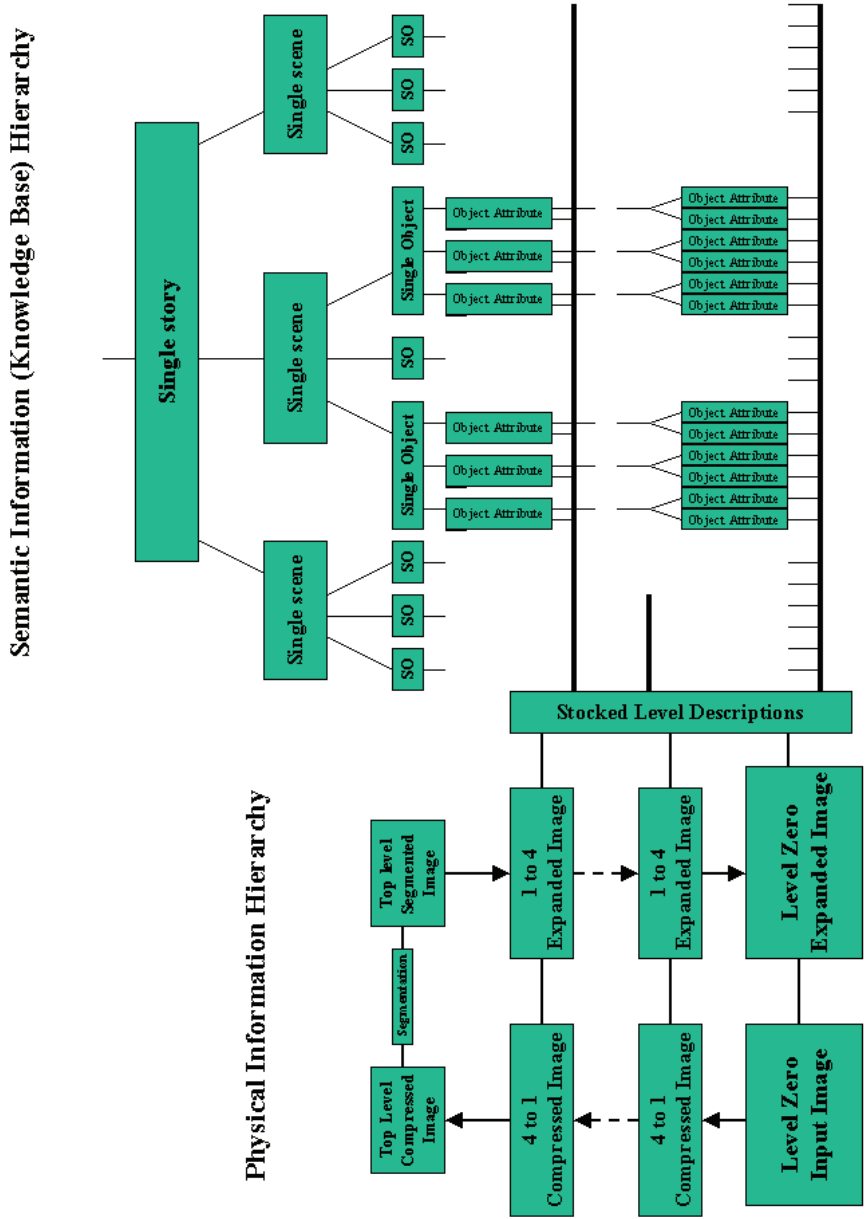

Figure 1. Arrangement of Physical and Semantic Information Hierarchies and their interconnection

Human's cognitive abilities (including the aptness for image interpretation and the capacity to assign semantics to an image) are empowered by the existence of a huge knowledge base about the things in the surrounding world kept in human brain/head.

This knowledge base is permanently upgraded and updated during the human's life span. So, if we intend to endow our visual robot with some cognitive capabilities we have to provide it with something equivalent to this (human) knowledge base.

It goes without saying that this knowledge base will never be as large and developed as its human prototype. But we are not sure that such a requirement is valid in our case. After all, humans are also not equal in their cognitive capacities, and the content of their knowledge bases is very diversified too. (The knowledge base of aerial photographs interpreter is certainly different from the knowledge base of X-ray images interpreter, or IVUS images, or PET images). The knowledge base of our visual robot has to be small enough to be effective and manageable, but sufficiently large to ensure the robot's acceptable performance. Certainly, for our feasibility study we can be satisfied even with a relatively small, specific-task-oriented knowledge base.

The next crucial point is the knowledgebase representation issue. To deal with it, we first of all must arrive at a common agreement about what is the meaning of the term "knowledge". (A question that usually has no commonly accepted answer.) We state that in our case a suitable and a sufficient definition of it would be: "**Kownledge is a memorized information**". Consequently, we can say that knowledge (like information) must be a hierarchy of descriptive items, with the grade of description details growing in a top-down manner at the descending levels of the hierarchy.

What else must be mentioned here, is that these descriptions have to be implemented in some alphabet (as it is in the case of physical information) or in a description language (which better fits the semantic information case). Any farther argument being put aside, we will declare that the most suitable language in our case is the natural human language. After all, the real knowledge bases that we are familiar with are implemented in natural human languages.

The next step, then, is predetermined: if natural language is a suitable description implement, the suitable form of this implementation is a narrative, a story tale (Tuffield et al., 2005). If the description hierarchy can be seen as an inverted tree, then the branches of this tree are the stories that encapsulate human's experience with the surrounding world. And the leaves of these branches are single words (single objects) from which the story parts (single scenes) are composed of.

The descent into description details, however, does not stop here, and each single word (single object) can be farther decomposed into its attributes and rules that describe the relations between the attributes.

At this stage the physical information reappears. Because the words are usually associated with physical objects in the real world, words' attributes must be seen as memorized physical information (descriptions). Once derived (by the HVS) from the observable world and learned to be associated with a particular word, these physical information descriptions are soldered in into the knowledgebase. Object recognition, thus, turns out to be a comparison and similarity test between currently acquired physical information and the one already retained in the memory. If the similarity test is successful, starting from this point in the hierarchy and climbing back up on the knowledgebase ladder we will obtain: first, the linguistic label for a recognized object; second, the position of this label (word) in the context of the whole story; and third, the ability to verify the validity of an initial guess by testing the appropriateness of the neighboring parts composing the object or the context of a story. In this way, object's meaningful categorization can be reached, and the first stage of image annotation can be successfully accomplished, providing the basis for farther meaningful (semantic) image interpretation.

One question has remained untouched in our discourse: How this artificial knowledgebase has to be initially created and brought into the robot's disposal? The vigilant reader certainly remembers the fierce debates about learning capabilities of neural networks and other machine learning technologies. We are aware of these debates. But in our case we can state certainly: they are irrelevant. For a simple reason: the top-down fashion of the knowledge base development pre-determines that all responsibilities for knowledge base creation have to be placed on the shoulders of the robot designer.

Such an unexpected twist in design philosophy will be less surprising if we recall that human cognitive memory is also often defined as a "declarative memory". And the prime mode of human learning is the declarative learning mode, when the new knowledge is explicitly transferred to a developing human from his external surrounding: From a father to a child, from a teacher to a student, from an instructor to a trainee. So, our proposal that robot's knowledgebase has to be designed and created by the robot supervisor is sufficiently correct and is fitting our general concept of information use and management.

## 5. More explanation is required

The proposed Semantic Information Processing scheme must be so annoyingly different from other knowledge-management forms that farther explanations in its defence must be provided. The vigilant reader has certainly paid attention to the fact that the term "ontology" does not appear in the text, albeit ontology is a ubiquitously used technique for human knowledgebase creation and representation. I have chose to avoid the use of the term ontology for the following reason.

More than twenty years ago, a famous Soviet mathematician, Israel Gelfand, and his colleagues were trying to devise a knowledge-based system for medical diagnostic problem solving. From the very beginning, the need for an adequate description language has become apparent, and extensive research efforts were spent moving toward this goal. The notion of ontology has not been yet known – the seminal paper of Thomas Gruber would appear only in the year 1993 (Gruber, 1993). However, in the preface to the book that summarizes their experience, which was well ahead of their time, while referring to the language creation difficulties, Israel Gelfand writes: "There are two ways to create a language: to compose literature scripts or to compile a dictionary. We all know how significant for the Russian language were the works of Pushkin and Dhale" (Gelfand et al., 1989). (We would add accordingly – Shakespeare and Dr. Johnson, for the English language).

The problem is that in contemporary knowledge-based systems design, ontology is used in only one of its manifestations – a vocabulary, a thesaurus. That is, certainly, a miss and a fault, which in our design we are trying to avoid. Imagine a Martian guest that is trying to understand our world relying only on the Oxford Concise Dictionary. On the other hand, you can easily recall the picture books with stories that the grandmother has read to you again and again in the childhood.

The story telling approach that we decided to pursue (and are trying to implement) is also very different from those that could be found in today's research papers. Current trend in story telling research and development is focused on automatic narrative creation, very similar to what is going on in the classical ontology design practice. In this regard it would be a proper place to remind that we reject the tradition of autonomous ontology creation. We are inclined to the "grandmother approach", where, as it was already explained earlier, the new knowledge comes to its possessor from the outside, from someone who already possesses it: A grandmother telling the child her stories, dancing bees that convey to the rest of the hive the information about melliferous sites (Zhang et al., 2005), ants that learn in tandem (Franks & Richardson, 2006), and even bacteria developing their antibiotic resistance as a result of a so-called horizontal gene transfer when a single DNA fragment of one bacteria is disseminated among other colony members (Lawrence & Hendrickson, 2003). That is, in our case this is a job for the robot's designer. In a story telling manner he has to transfer to the robot his view on the surrounding world and his understanding of a proper behavior in different task-inspired situations.

I am aware that by denying the bottom-up machine-learning-inspired knowledge acquisition I am awaking all the bears in my environment. But sorry, that is only an attempt to find out the way to leave the dead-ended alley where image processing is stalled for so many years.

Let us continue: Vigilant readers have certainly also paid attention to the fact that the name of Claude Shannon (the famous inventor of the Information Theory of Communication) is not mentioned in the paper. The reason for this is clear and plain – Shannon says nothing about the notion of information, about "What is information?" He has invented a measure of information, but that says nothing about the notion of information. Like the measure of time, which we ubiquitously use (second, hour, day, etc.) tells nothing about the notion of time, about "What is time?".

Kolmogorov too was busy with very different things. Randomness has been his main concern. According to the Kolmogorov's theory, a message composed as a sequence of random values cannot be depicted (reproduced) by a description program, which is shorter than the original message. That is, the description of a random message is the message itself. What follows from this, is that nonrandom data structures could be described in a concise compressed form, which Chaitin calls "Algorithmic Information" (Chaitin, 1977), Floridi – "Meaningful data" (Floridi, 2005), Vitanyi – "Meaningful Information" (Vitanyi, 2006). That means that each message can be seen as a composition of: a compressible, information-bearing part of it and a non-compressible, information-devoid, random data part. The first part we call Physical Information, and it is obvious that processing only this part of the message will give us a tremendous gain against the data processing case where meaningful and meaning-less data are inseparable.

The March 2008 issue of the IEEE Signal Processing Magazine is entirely devoted to this problem: in different domains of signal processing people have empirically discovered the advantages of what they call "Compressive Sampling". In the preface to the magazine the guest editors write: "At the heart of the new approach are two crucial observations. The first is that the Shannon/Nyquist signal representation exploits only minimal prior knowledge about the signal being sampled, namely its bandwidth. However, most objects we are interested in acquiring are structured and depend upon a small number of degrees of freedom than the bandwidth suggests. In other words, most objects of interest are sparse or compressible in the sense that they can be encoded with just a few numbers without numerical or perceptual loss". Bravo! There could be no better explanation to the benefits of information processing versus brute force data processing. The tradition is, however, stronger than the reason – the rest of the magazine is devoted to the alchemy of compressive sampling accomplishment via bottom-up raw data processing.

Some words I would like to spend on the latest developments in the HVS research. While the mainstream of human vision research continues to approach visual information processing in a bottom-up feed-forward fashion (Serre et al., 2005; Kveraga et al., 2007) it turns out that the idea of primary top-down processing was never extraneous to biological vision. The first publications addressing this issue are dated by the early eighties of the last century, (Navon, 1977; Chen, 1982). The prominent authors were persistent in their claims, and farther research reports were published regularly until the recent time, (Navon, 2003; Chen, 2005). However, it looks like they have been overlooked, both in biological and in computer vision research. Only in the last years, a tide of new evidence has become visible and is pervasively discussed now. Although the spirit of these discussions is still different from our view on the subject, the trend is certainly in favor of the foremost top-down visual information processing (Ahissar & Hochstein, 2004; Juan et al., 2004). Again, top-down information processing in the physical information processing part only is assumed here. Information processing partition proposed in this paper is not acknowledged by the contemporary vision researchers.

## 6. Some conclusions

In this paper, I have proposed a few ideas that are entirely new and therefore might look suspicious. All the novelties come as a natural extension of a new definition of information that is sequentially applied to various aspects of image processing. The most important innovation is positing information image processing as the prime mode of image processing (in contrast to traditionally dominant data image processing). The next novelty is the dissociation between physical and semantic information processing within the information-processing domain. The proposed arrangement of information-processing hierarchies is a further extension of the basic idea of the information-processing nature of the HVS, and its imitation in an artificial vision system – our hypothetical visual robot design.

Despite of the skeptical welcome, the efficiency of the unsupervised top-down directed region-based image segmentation is hard to disprove today. Although the story telling approach to knowledgebase hierarchy creation is not yet so rigorously proved, we hope that this development stage will also be successfully surmounted.

I hope that the time of our persuasive success is not far away.